\documentclass[lettersize,journal,]{IEEEtran}
\usepackage{amsmath,amsfonts,amssymb}
\usepackage{booktabs,makecell,xcolor}
\usepackage{algorithmic}
\usepackage{algorithm}
\usepackage{array}
\usepackage[caption=false,font=normalsize,labelfont=sf,textfont=sf]{subfig}
\usepackage{textcomp}
\usepackage{stfloats}
\usepackage{url}
\usepackage{verbatim}
\usepackage{graphicx}
\graphicspath{{figs/}}
\usepackage{cite}
\begin{document}

\title{V2X-DSC: Multi-Agent Collaborative Perception with Distributed Source Coding Guided Communication}


 \author{
Yuankun Zeng, 
Shaohui Li, 
Zhi Li, 
Shulan Ruan,
Yu Liu, 
and You He
\thanks{
Yuankun Zeng, Zhi Li, and Shulan Ruan are with Shenzhen International Graduate School, Tsinghua University, Shenzhen 518055, China (e-mail: 
zengyk24@mails.tsinghua.edu.cn; zhilizl@sz.tsinghua.edu.cn; slruan@sz.tsinghua.edu.cn).}% 
\thanks{Shaohui Li is with College of Information Science and Electronic Engineering, Zhejiang University, Hangzhou 310007, China (e-mail: lishaohui@zju.edu.cn).}% 
\thanks{Yu Liu and You He are with Department of Electronic Engineering, Tsinghua University, Beijing 10084, China (e-mail: liuyu\_thu@mail.tsinghua.edu.cn; 
heyou@mail.tsinghua.edu.cn).}
}


% % column for its text to clear the IEEEpubid mark.

\maketitle

\begin{abstract}
Collaborative perception improves 3D scene understanding by fusing complementary observations from multiple connected agents. 
Despite impressive gains, intermediate-feature sharing remains difficult to deploy due to strict communication budgets: dense BEV feature maps are high-dimensional and saturate practical V2X links even after spatial selection.
A key but under-exploited fact is that collaborators observe the same physical world and thus their intermediate features are strongly correlated.
This suggests that a receiver should not require a full description of a collaborator feature; instead, it should only need the innovation that is not predictable from its own local representation. In this paper, we revisit collaborative perception from a distributed source coding perspective and propose V2X-DSC, a DSC-guided framework with a DSC-guided Conditional Codec (DCC) for bandwidth-constrained intermediate fusion. For each directed interaction, the sender encodes pruned BEV features into discrete symbols via vector quantization and packs them into a compact bitstream using rANS entropy coding. The receiver performs conditional reconstruction by exploiting its local BEV feature as side information, which provides rich scene context and semantic priors. This context allows the decoder to infer predictable content and allocate the transmitted bits mainly to complementary cues, yielding a high-fidelity representation optimized for downstream fusion under tight bandwidth. Crucially, this conditional framework imposes a structural regularization on the learning process: the encoder is encouraged to represent incremental information beyond the receiver context, while the reconstruction is constrained by the side-information prior, resulting in lower-noise and more consistent features.
Experiments on DAIR-V2X, OPV2V, and V2X-Real show that V2X-DSC achieves state-of-the-art accuracy--bandwidth trade-offs under \textbf{KB-level} per-link communication, and generalizes as a plug-and-play communication layer across multiple fusion backbones.
\end{abstract}


\section{Introduction}

Collaborative perception enables connected agents to overcome the inherent limitations of single-view sensing by aggregating complementary observations from multiple viewpoints~\cite{wang2020v2vnet, xu2022opv2v}. By fusing multi-agent cues, collaborative perception can mitigate occlusions, extend sensing range, and enhance scene understanding in complex traffic scenarios. Among alternative collaboration mechanisms, intermediate feature fusion has emerged as a dominant design due to its favorable accuracy and modularity: each agent extracts a BEV feature map with a backbone network and exchanges intermediate representations for downstream fusion.

Despite its effectiveness, intermediate fusion faces a fundamental obstacle in practice: communication. BEV feature maps are spatially dense and high-dimensional, making frequent multi-agent exchange prohibitively expensive under realistic V2X bandwidth budgets. This bottleneck is further amplified by multi-neighbor collaboration, where limited channel capacity must be shared among multiple links. As a result, practical deployment requires communication to be reduced to the kilobyte scale per directed exchange, without sacrificing task performance.

To address this challenge, existing work primarily follows two directions. The first performs critical information selection, transmitting only salient agents or spatial regions predicted to benefit perception~\cite{liu2020who2com, liu2020when2com, hu2022where2comm, xu2025cosdh, tang2025cost}. The second develops compact feature representations via learned compression~\cite{hu2024communication, ding2025point, liu2025sparsecomm}. While successful, most approaches treat a collaborator feature as a standalone source to be conveyed, focusing on reducing the marginal complexity of what is transmitted. This overlooks a fundamental premise of collaborative perception: inter-agent correlation. Nearby agents observe the same physical world and encode it into an aligned latent space, inducing strong statistical dependence between their intermediate features. Consequently, transmitting a complete description of a collaborator feature is inherently redundant-the receiver's local feature already provides a strong prior over the shared scene context, so bits are better spent on complementary cues.

\textbf{We have noticed that DPFC~\cite{11358789} also employs a DSC framework for compressing Lidar point cloud feature for autonomous driving, which has a insight similar to our work. We would like to clarify that these two studies are concurrent works and developed from different backbones. Thus, the detailed designs and experimental results are distinct.}

This paper leverages this overlooked opportunity through the lens of Distributed Source Coding (DSC). DSC theory establishes that correlated sources can be compressed efficiently with separate encoders, provided the decoder can exploit correlation through jointly available context. In collaborative perception, the receiver's local BEV feature naturally serves as such context. This perspective suggests an innovation-centric communication objective: a sender should transmit primarily the complementary information that is not predictable from the receiver-local representation, rather than a self-contained copy of its feature.

Based on this insight, we propose V2X-DSC, a DSC-guided collaborative perception framework equipped with a DSC-guided Conditional Codec (DCC) for bandwidth-constrained intermediate fusion. For each directed interaction, the sender first prunes task-irrelevant BEV regions to avoid spending bandwidth on free space, then maps the remaining feature into discrete symbols via vector quantization and packs them into a compact bitstream using rANS entropy coding. At the receiver, the bitstream is decoded losslessly back to indices and dequantized. Crucially, reconstruction is performed conditionally: the receiver utilizes a side-information network to extract context from its local BEV feature, providing rich geometric and semantic priors. This context allows the decoder to infer predictable content and allocate the transmitted bits mainly to complementary cues, yielding a high-fidelity representation optimized for downstream fusion under tight bandwidth. Furthermore, this conditional framework imposes a structural regularization on learning: the encoder is encouraged to represent incremental information beyond the receiver context, while reconstruction is constrained by the side-information prior, resulting in lower-noise and more fusion-consistent features.

We validate V2X-DSC on DAIR-V2X~\cite{yu2022dair}, OPV2V~\cite{xu2022opv2v}, and V2X-Real~\cite{xiang2024v2xreal}. Experiments demonstrate that our approach achieves state-of-the-art accuracy--bandwidth trade-offs and serves as a plug-and-play communication layer across multiple fusion backbones. We further evaluate robustness under pose noise and communication delay, showing stable performance under common system imperfections.

Our main contributions are summarized as follows:
\begin{itemize}
    \item We revisit bandwidth-constrained collaborative perception through the lens of Distributed Source Coding (DSC). By identifying inter-agent feature correlation as a theoretical basis for compression, we propose an innovation-centric communication paradigm that eliminates the redundancy of transmitting shared context.
    \item We propose V2X-DSC with a DSC-guided Conditional Codec (DCC) that follows an asymmetric design: the sender independently encodes its feature, while the receiver performs conditional decoding jointly with its local feature. By exploiting inter-agent feature correlation, DCC reduces the required bandwidth and yields more fusion-consistent, lower-noise collaborator features in practice.
    \item Extensive evaluations on OPV2V, DAIR-V2X, and V2X-Real demonstrate that our method establishes a new Pareto frontier, achieving kilobyte-level per-link communication while matching or surpassing strong baselines.
\end{itemize}

\section{Related Works}
\subsection{Collaborative Perception}
Perception capabilities constitute the cornerstone of autonomous agents. While single-agent systems face inherent physical limitations such as occlusion and long-range data sparsity, Multi-Agent Collaborative Perception mitigates these bottlenecks by facilitating information sharing among connected agents~\cite{xu2022opv2v,chen2019fcooper,wang2020v2vnet,li2021disconet}. 

However, transitioning from ideal simulations to real-world deployment faces severe practical impediments. The effectiveness of collaboration is strictly bounded by communication constraints, where limited bandwidth and packet loss necessitate highly efficient transmission strategies~\cite{hu2022where2comm,liu2020who2com,wang2020v2vnet}. Furthermore, system imperfections such as asynchronous latency and localization noise inevitably degrade fusion alignment~\cite{lei2022latency}. 
To counteract these physical errors, methods such as SyncNet~\cite{lei2022latency} and CoBEVFlow~\cite{wei2023cobevflow} introduced feature-level compensation and flow-based motion prediction, while CoDynTrust~\cite{xu2025codyntrust} and MRCNet~\cite{hong2024multi} further advanced robustness by explicitly modeling uncertainty and motion dynamics. 
Beyond physical constraints, addressing the challenge of agent heterogeneity, frameworks like HEAL~\cite{lu2024heal} and STAMP~\cite{gaostamp} were proposed to enable seamless collaboration across diverse sensor configurations and model architectures without systemic retraining. 
Recognizing the vulnerability of open communication networks, recent works such as CP-Guard~\cite{hu2025cp} and ROBOSAC~\cite{li2023robosac} have focused on \textbf{security}, developing mechanisms to detect malicious agents and defend against adversarial attacks.

\subsection{Communication Efficient Collaborative Perception}
Among the myriad challenges facing real-world deployment, communication-constraint stands as the most fundamental bottleneck, serving as the physical bedrock upon which all collaborative mechanisms are built. Practical networks (e.g., DSRC, C-V2X) operate under strict bandwidth budgets and latency limits. The direct transmission of high-dimensional sensory data or even dense intermediate features can rapidly saturate network capacity, inducing prohibitive delays that undermine safety. 

To reconcile high-bandwidth demands with limited network capacity, existing research has bifurcated into two dominant strategies: critical information selection and compact feature representation.

The first stream focuses on information selection, premised on the insight that not all spatial regions or feature channels contribute equally to perception tasks. To eliminate redundancy, methods like Who2com~\cite{liu2020who2com} and When2com~\cite{liu2020when2com} employ handshake mechanisms with attention weights to selectively share information only when it offers a distinct gain. Building on this, Where2comm~\cite{hu2022where2comm} utilizes spatial confidence maps to identify and transmit only perceptually critical regions. CoST~\cite{tang2025cost} adopts a unified spatio-temporal perspective to avoid re-transmitting static background features, while CoSDH ~\cite{xu2025cosdh} introduces a supply-demand awareness mechanism, ensuring transmission occurs only when the sender's high-confidence regions match the receiver's blind spots.

Complementary to selection, the second stream pursues compact feature representation through advanced compression techniques. These methods aim to encode the entire scene into a more efficient form. Early works like OPV2V ~\cite{xu2022opv2v} utilized simple autoencoders, while V2X-ViT ~\cite{xu2022v2x} employed $1\times1$ convolutions for channel reduction. Recent research has moved towards discrete and sparse representations. Codefilling~\cite{hu2024communication} discretizes continuous features into codebook indices, transforming transmission into a lookup task. CPPC~\cite{ding2025point} and SparseComm~\cite{liu2025sparsecomm} abandon dense grids in favor of sparse vector sets or clustered points to represent objects.

Despite these advances, existing approaches implicitly treat each agent's feature as an independent source to be compressed or selected, overlooking the statistical correlation between collaborator and ego features. This inter-agent redundancy suggests that transmitting full feature content---even in compressed form---is inefficient; the sender should instead convey only the innovation beyond what the receiver can infer.

\subsection{Distributed Source Coding}

Distributed Source Coding (DSC) provides the information-theoretic foundation for compressing correlated sources that are encoded independently but decoded jointly. The theoretical bounds were originally established by the Slepian-Wolf theorem~\cite{slepian2003noiseless} for lossless compression and extended to the lossy regime by the Wyner-Ziv theorem~\cite{wyner2003rate}. Practical implementations of DSC have historically relied on channel coding techniques, with prominent examples based on turbo codes~\cite{aaron2002compression} and LDPC codes~\cite{liveris2002compression}. These methods found success in distributed video coding, exemplified by the DISCOVER codec~\cite{artigas2007discover}.

Recently, the field has begun to integrate deep learning to model complex, non-linear correlations that traditional channel codes struggle to capture. Emerging works have established Neural DSC frameworks~\cite{whang2024neural} that approximate Wyner-Ziv limits using generative networks. Learning-based distributed image codecs have been proposed for multi-view~\cite{zhangldmic, huang2023learned} and stereo~\cite{ayzik2020deep, xia2025fca} settings. These ideas have also been applied to deep video compression~\cite{zhang2023low}.

\section{Problem Formulation and Background}
\label{sec:problem_background}

In this section, we formulate bandwidth-constrained collaborative perception and summarize the distributed source coding (DSC) principles that motivate our design.
\subsection{Collaborative Perception under Bandwidth Constraints}
\label{subsec:problem_formulation}

We consider a collaborative perception system with $N$ agents.
At each time step, agent $k \in \{1,\ldots,N\}$ acquires sensor observations $\mathcal{O}_k$ and extracts an intermediate bird's-eye-view (BEV) feature map
$\mathbf{F}_k \in \mathbb{R}^{C \times H \times W}$ using a backbone network:
\begin{equation}
\mathbf{F}_k = \Phi(\mathcal{O}_k).
\end{equation}
Collaboration aims to improve a downstream task, such as 3D detection, by exchanging compact messages derived from these intermediate representations.

Without loss of generality, we describe prediction from the perspective of an arbitrary agent $i$.
Let $\mathcal{N}(i)$ denote the set of neighboring agents that send messages to $i$ at the current time step.
Each neighbor $j \in \mathcal{N}(i)$ transmits a message $m_{j\rightarrow i}$, which is decoded into a fusion-ready representation $\hat{\mathbf{F}}_{j\rightarrow i}$.
Agent $i$ then fuses its local feature with the decoded representations and produces the task prediction:
\begin{equation}
\hat{\mathbf{y}}_i =
\mathrm{Head}\Big(\mathrm{Fuse}\big(\mathbf{F}_i,\{\hat{\mathbf{F}}_{j\rightarrow i}\}_{j\in\mathcal{N}(i)}\big)\Big).
\end{equation}
We seek to optimize task performance under stringent communication budgets. Let $\Theta$ denote the learnable parameters of the collaborative perception model.
Our objective is to minimize the expected task loss subject to a per-message bandwidth budget:
% \begin{equation}
% \min_{\Theta} \ \mathbb{E}\big[\mathcal{L}_{task}(\hat{\mathbf{y}}_i,\mathbf{y}^{gt}_i)\big]
% \quad \text{s.t.} \quad |m_{j\rightarrow i}| \leq B,\ \forall\, j\in\mathcal{N}(i),
% \label{eq:optimization}
% \end{equation}
\begin{equation}
\begin{split}
& \min_{\Theta} \ \mathbb{E}\big[\mathcal{L}_{task}(\hat{\mathbf{y}}_i,\mathbf{y}^{gt}_i)\big] \\
& \text{s.t.} \quad |m_{j\rightarrow i}| \leq B,\ \forall\, j\in\mathcal{N}(i),
\end{split}
\label{eq:optimization}
\end{equation}
where $B$ is the bandwidth budget for each exchanged message.

\subsection{Communication Bottleneck Analysis}
\label{subsec:bottleneck}

Intermediate feature sharing provides strong performance in collaborative perception, but quickly becomes communication-prohibitive under realistic bandwidth constraints.
Let $\mathbf{F}\in\mathbb{R}^{C\times H\times W}$ denote a typical intermediate BEV feature map to be exchanged between agents.
If transmitted in raw form with $b$ bits per scalar, the per-frame payload is
\begin{equation}
R_{\text{raw}} = C \cdot H \cdot W \cdot b \quad \text{(bits/frame)}.
\end{equation}
For representative BEV features (e.g., $\mathbf{F}\in\mathbb{R}^{64\times 256\times 256}$ with $b=32$), this corresponds to approximately $16$ MB per frame.
At a typical perception frequency of $10$ Hz, the sustained data rate per agent exceeds $1$ Gbps, i.e., orders of magnitude beyond the capacity of vehicular communication technologies such as DSRC ($\sim$10 Mbps) or C-V2X sidelink ($\sim$50--100 Mbps shared among neighbors).

Consequently, collaborative perception demands aggressive compression in practice, often requiring $100$--$1000\times$ reduction to fit within kilobyte-scale budgets.
Existing methods achieve this via spatial downsampling, channel reduction, or learned quantization.
However, these approaches treat each agent's feature as an independent source, compressing $\mathbf{F}_j$ without exploiting its statistical correlation with the receiver's local observation $\mathbf{F}_i$.
Because collaborating agents observe spatially proximate or semantically related regions, their features exhibit substantial redundancy—information that independent coding fails to leverage.

This observation motivates a shift toward conditional coding, where compression exploits the fact that the decoder possesses correlated side information.

\subsection{Distributed Source Coding Theorem}
\label{subsec:dsc_background}

Distributed source coding studies how to compress correlated sources with separate encoders and a joint decoder that exploits their statistical dependence. Its key insight is that, for correlated sources, communication cost can be reduced by leveraging statistical dependence at the decoder rather than transmitting each source independently.

\paragraph{Slepian--Wolf theorem (lossless DSC).}
Consider two correlated discrete memoryless sources $(X,Y)$.
Even if $X$ and $Y$ are encoded independently, there exist coding schemes such that a joint decoder can recover both sources losslessly as long as the encoding rates satisfy the Slepian--Wolf region:
\begin{equation}
\begin{gathered}
R_X \ge H(X|Y), \\
R_Y \ge H(Y|X), \\
R_X + R_Y \ge H(X,Y).
\end{gathered}
\label{eq:slepian_wolf}
\end{equation}
This result implies that separate encoding can theoretically achieve the efficiency of joint encoding.
Crucially, if the decoder already has access to $Y$ as \textit{Side Information}, the source $X$ can be compressed to its conditional entropy $H(X|Y)$. Since $H(X|Y) = H(X) - I(X;Y)$, the term $I(X;Y)$ represents the theoretical bandwidth saving achievable by exploiting source correlations.

\paragraph{Wyner--Ziv Coding (Lossy DSC with Side Information).}
The Wyner--Ziv theorem extends DSC to the lossy regime, addressing the compression of a source $X$ when correlated side information $Y$ is available strictly at the decoder.
The encoder maps $X$ to a compressed representation, while the decoder leverages $Y$ to produce a reconstruction $\hat{X}$ satisfying an expected distortion constraint $\mathbb{E}[d(X,\hat{X})]\le D$.
The operational rate--distortion function is characterized by:
\begin{equation}
R_{WZ}(D) = \inf_{p(z|x)} I(X;Z\,|\,Y),
\label{eq:wyner_ziv}
\end{equation}
where $Z$ is an auxiliary random variable forming a Markov chain $Y \leftrightarrow X \leftrightarrow Z$. This Markov structure mathematically enforces the constraint that the encoding of $X$ must be independent of $Y$.
Crucially, the theorem proves that $R_{WZ}(D) \le R_{X}(D)$, where $R_{X}(D)$ is the standard rate-distortion function without side information.
This inequality provides the theoretical foundation for our framework: it implies that by exploiting the correlation structure at the decoder, the communication burden is no longer governed by the intrinsic complexity of $X$, but rather by the conditional innovation of $X$ given $Y$.

\subsection{Inter-Agent Correlation: A DSC Perspective}
\label{subsec:connection}

We now connect the DSC framework to bandwidth-constrained collaborative perception.
Consider communication from a collaborating agent to a receiving agent.
Let $X$ denote the transmitted intermediate representation and $Y$ denote the receiver's locally available representation at decoding time.
In our setting, we instantiate this correspondence as
\begin{equation}
X \ \triangleq\  \mathbf{F}_{j}, \qquad
Y \ \triangleq\  \mathbf{F}_{i},
\end{equation}
where $\mathbf{F}_{j}$ is the collaborator's feature to be communicated and $\mathbf{F}_{i}$ is the receiving agent's own feature extracted from its local observations.

\subsubsection{The Overlooked Opportunity: Inter-Agent Correlation}
A fundamental yet underutilized property of collaborative perception is that intermediate representations across agents exhibit strong statistical dependence: nearby agents observe the same physical environment and employ similarly trained encoders.
This dependence enables conditional communication, when the receiver already possesses informative scene context, transmitting a complete description of the collaborator's feature becomes redundant.
We identify three principal sources of inter-agent correlation in collaborative perception scenarios:

\paragraph{Overlapping spatial observations.}
Agents operating in proximity frequently have partially overlapping sensor coverage, particularly in dense traffic or structured environments such as intersections.
Co-visible objects and scene elements induce semantically and geometrically aligned cues across agents, resulting in correlated feature map activations.

\paragraph{Shared scene context.}
Even when observable regions differ, agents operate within a common scene and share contextual priors: road topology, lane structure, static infrastructure, and traffic patterns.
Such global scene structure is implicitly encoded in deep BEV representations, inducing correlation beyond direct field-of-view overlap.

\paragraph{Aligned representation learning.}
Deep perception backbones are trained to map heterogeneous sensor inputs into a task-optimized latent space.
Through end-to-end learning, semantically similar entities produce consistent latent patterns across viewpoints, reducing representational discrepancy and enhancing cross-agent predictability.

\subsubsection{Implications and Challenges}
From the DSC perspective, the above correlations imply that the communication cost for transmitting the collaborator's feature should be governed by its conditional information content given the receiver's local representation, rather than its marginal complexity alone.
This motivates designing codecs that convey primarily the complementary information in $\mathbf{F}_{j}$ beyond what is already captured by $\mathbf{F}_{i}$.

However, operationalizing DSC theory in collaborative perception is nontrivial.
The sources are high-dimensional continuous feature maps; the objective is downstream task performance under strict bandwidth constraints; and practical systems require end-to-end trainable modules that integrate seamlessly with modern fusion architectures.
In the next section, we present a DSC-guided neural codec framework that realizes this conditional communication principle for collaborative perception.

\section{Methods}

 \begin{figure*}[t]
  \centering
  \includegraphics[width=\textwidth]{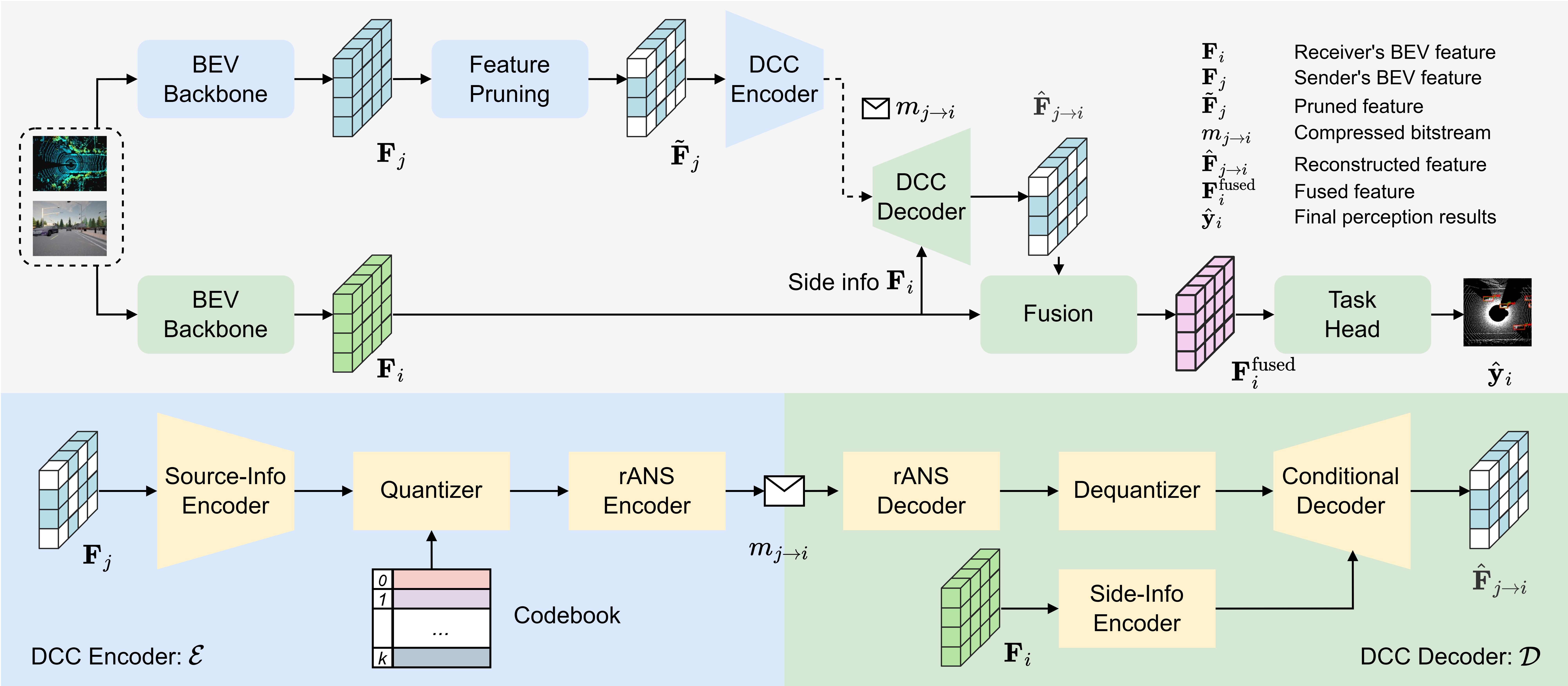}
    \caption{\textbf{Overall pipeline of V2X-DSC with the proposed DSC-guided Conditional Codec (DCC).}
    Each agent extracts a BEV feature map $\mathbf{F}_k=\Phi(\mathcal{O}_k)$ from its local observation.
    For a directed interaction $j\rightarrow i$, the sender prunes task-irrelevant regions to obtain $\tilde{\mathbf{F}}_j$, encodes it into a compressed bitstream $m_{j\rightarrow i}$, and transmits it to the receiver.
    The receiver decodes $m_{j\rightarrow i}$ and reconstructs a fusion-ready feature $\hat{\mathbf{F}}_{j\rightarrow i}$ conditioned on its local feature $\mathbf{F}_i$ (side information).
    Reconstructed features are fused with $\mathbf{F}_i$ to form $\mathbf{F}^{\mathrm{fused}}_i$, which is fed to the task head to produce the final prediction $\hat{\mathbf{y}}_i$.
    \textbf{Color coding:} blue indicates sender-side processing and transmitted representations, while green indicates receiver-side processing that leverages locally available information.}
    \label{fig:pipeline}
  \end{figure*}

\subsection{Modeling Collaborative Perception as Distributed Coding}
\label{subsec:modeling_dsc}

We model bandwidth-constrained collaboration as a distributed coding problem in which the receiver can exploit locally available correlated context.
Consider an arbitrary receiving agent $i$ and one of its neighbors $j\in\mathcal{N}(i)$.
Both agents extract intermediate BEV feature maps $\mathbf{F}_i,\mathbf{F}_j \in \mathbb{R}^{C\times H\times W}$ from their local observations.
The goal of communication on the directed interaction $j\rightarrow i$ is to convey a compact description of $\mathbf{F}_j$ that remains useful for fusion and downstream prediction at agent $i$ under a strict bandwidth budget.

\paragraph{Source--side information mapping.}
We map collaborative perception variables to the standard DSC formulation:
\begin{equation}
X \triangleq \mathbf{F}_j, \qquad Y \triangleq \mathbf{F}_i.
\end{equation}
Due to inter-agent correlation induced by shared scene context and partially overlapping visibility, $X$ and $Y$ are statistically dependent under the data distribution.
Consistent with the Wyner--Ziv setting, encoding is performed \emph{independently} in the sense that the sender compresses $X$ without access to the realization of $Y$.
Decoding, however, can be \emph{context-aware}: the receiver leverages its locally available representation $Y$ when interpreting the received message and forming a reconstruction for fusion.

\paragraph{Budget-constrained conditional reconstruction.}
Let $\mathcal{E}$ and $\mathcal{D}$ denote the sender-side encoder and receiver-side decoder on link $j\rightarrow i$.
The sender forms a message
\begin{equation}
m_{j\rightarrow i} = \mathcal{E}(\mathbf{F}_j),
\end{equation}
and the receiver reconstructs a fusion-ready representation by combining the message with its local feature:
\begin{equation}
\hat{\mathbf{F}}_{j\rightarrow i} = \mathcal{D}(m_{j\rightarrow i}, \mathbf{F}_i).
\end{equation}
We seek to minimize downstream task loss subject to a per-message budget $|m_{j\rightarrow i}|\le B$.
The DSC principle suggests that, when $Y$ is informative, the message need not convey the full marginal content of $X$; instead, it should prioritize the information in $X$ that is complementary to $Y$, corresponding to the residual uncertainty of $X$ conditioned on $Y$.

\paragraph{DSC-guided learning in high-dimensional feature spaces.}
Classical DSC constructions rely on structured codes and explicit probabilistic modeling.
In collaborative perception, the variables are high-dimensional continuous feature maps, making explicit modeling of the joint distribution of $(X,Y)$ impractical.
We therefore adopt a DSC-guided approach that realizes conditional coding implicitly within a deep learning framework.
Rather than constructing explicit binning schemes, we learn an end-to-end communication module whose discrete bottleneck and conditional decoding mechanism encourage the model to suppress information in $\mathbf{F}_j$ that is redundant given $\mathbf{F}_i$, while preserving task-relevant complementary cues for fusion.

\subsection{Overall Pipeline}
\label{subsec:pipeline}

Figure~\ref{fig:pipeline} illustrates the overall framework of V2X-DSC.
We consider an intermediate-fusion collaborative perception pipeline in which agents exchange compact messages derived from BEV features under strict bandwidth constraints.
Our key component is the proposed DSC-guided Conditional Codec (DCC), which compresses collaborator features into a bitstream and enables conditional reconstruction at the receiver.

\paragraph{BEV feature extraction.}
At each time step, each agent $k$ processes its local observation $\mathcal{O}_k$ using a BEV backbone $\Phi$ and obtains an intermediate feature map
$\mathbf{F}_k=\Phi(\mathcal{O}_k)\in\mathbb{R}^{C\times H\times W}$.
For a directed communication $j\rightarrow i$, $\mathbf{F}_j$ denotes the sender feature and $\mathbf{F}_i$ denotes the receiver-local feature.

\paragraph{Task-irrelevant pruning.}
Not all spatial locations in the BEV feature map are equally informative for downstream tasks.
Background regions often contribute marginally to detection yet consume bandwidth if transmitted~\cite{hu2022where2comm}.
We therefore apply a spatial pruning step before encoding.
Sender $j$ predicts an ROI score map $S_j \in [0,1]^{H \times W}$, and derives a binary transmission mask
$M_j = \mathbf{1}(S_j > \tau)$, where $\tau$ controls the trade-off between coverage and bandwidth.
The pruned feature is $\tilde{\mathbf{F}}_j = M_j \odot \mathbf{F}_j$, which retains informative regions for subsequent encoding.

\paragraph{DSC-guided communication.}
For each directed interaction $j\!\rightarrow\! i$, the sender encodes the pruned feature $\tilde{\mathbf{F}}_j$ independently using the DCC encoder $\mathcal{E}$.
The encoder outputs discrete symbols via vector quantization, which are further packed into a compact bitstream using rANS entropy coding, yielding the transmitted message $m_{j\rightarrow i}$.
At the receiver, rANS decoding and dequantization recover the discrete latents, and the DCC decoder $\mathcal{D}$ performs conditional reconstruction by leveraging the receiver-local feature $\mathbf{F}_i$ through a side-information encoder.
This conditioning provides rich scene context, allowing the decoder to infer predictable content and devote the transmitted bits to complementary cues that are most useful for fusion under tight bandwidth.

\paragraph{Multi-agent fusion and prediction.}
After decoding, the receiver aggregates its local feature with reconstructed features from all neighbors $j\in\mathcal{N}(i)$:
\begin{equation}
\mathbf{F}^{\mathrm{fused}}_i
=
\mathrm{Fuse}\Big(\mathbf{F}_i, \{\hat{\mathbf{F}}_{j \rightarrow i}\}_{j \in \mathcal{N}(i)}\Big).
\end{equation}
In our implementation, $\mathrm{Fuse}(\cdot)$ adopts an FPN-style multi-scale aggregation~\cite{lin2017feature, lu2024heal} with element-wise max fusion~\cite{chen2019fcooper} to accommodate a varying number of neighbors. Finally, a task head produces the 3D detection output $\hat{\mathbf{y}}_i = \mathrm{Head}(\mathbf{F}^{\mathrm{fused}}_i)$.

\subsection{DSC-guided Conditional Codec}
\label{subsec:dcc}
\begin{figure*}[t]
    \centering
    \includegraphics[width=\textwidth]{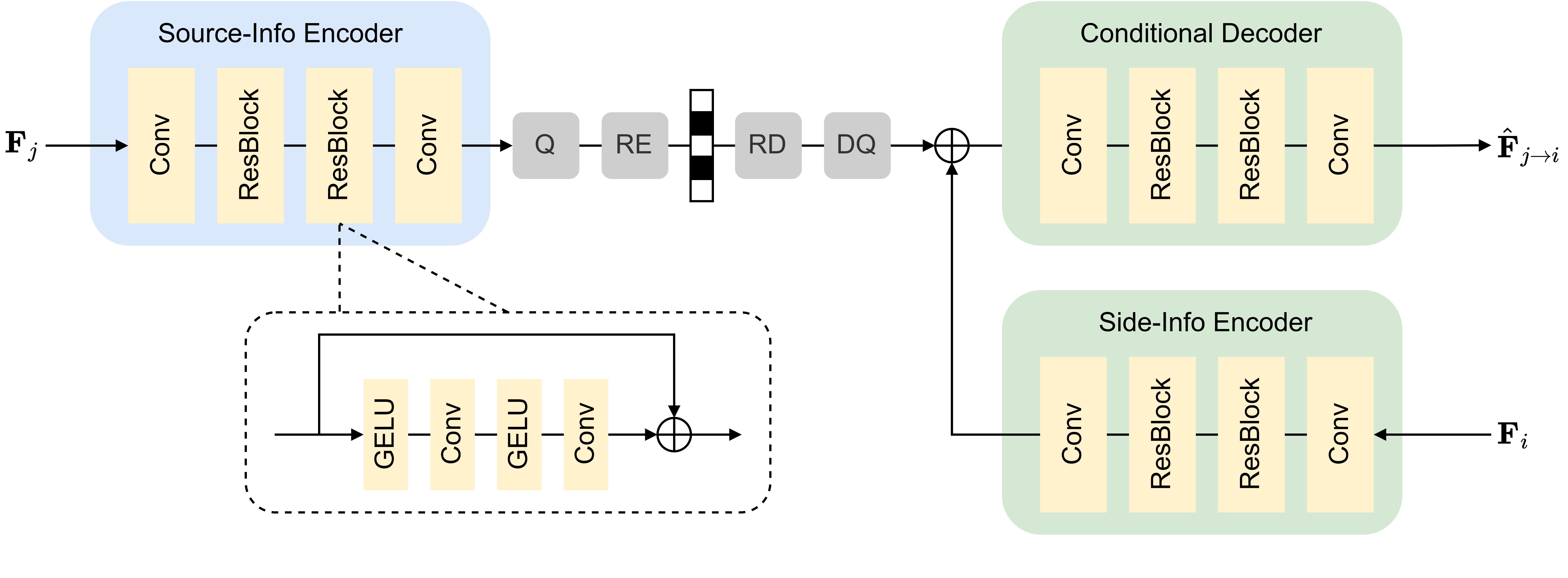}
    \caption{\textbf{Architecture of the DSC-guided Conditional Codec (DCC).}
    Given the sender feature $\mathbf{F}_j$, the source-information encoder maps it to a compact latent representation, which is discretized by codebook quantization (Q).
    The resulting discrete symbols are entropy-coded into a compressed bitstream via rANS encoding (RE) and transmitted over the communication channel.
    On the receiver side, rANS decoding (RD) and dequantization (DQ) recover the quantized latent.
    In parallel, the receiver-local feature $\mathbf{F}_i$ is processed by a side-information encoder to produce conditioning features.
    The conditional decoder then combines the decoded latent with the side-information branch and reconstructs a fusion-ready feature $\hat{\mathbf{F}}_{j\rightarrow i}$.}
    \label{fig:dcc}
\end{figure*}

Figure~\ref{fig:dcc} details the proposed DCC module. We detail the proposed DSC-guided codec used on each directed interaction $j\rightarrow i$.
The codec follows an asymmetric design: the sender compresses its pruned feature without access to the receiver feature, while the receiver performs conditional reconstruction using its locally available feature.
This realizes the DSC principle in a learnable manner and is compatible with end-to-end task training.

\paragraph{Sender-side DSC encoder.}
Given the pruned BEV feature $\tilde{\mathbf{F}}_j$, the sender applies a lightweight convolutional encoder to produce a compact latent representation.
The encoder is intentionally shallow and operates at the same spatial resolution as the BEV feature to preserve spatial correspondence.
Following common residual design, we optionally add a skip connection from the input feature to stabilize optimization and retain low-level structure.

\paragraph{Discrete bottleneck via codebook quantization.}
To enable kilobyte-level communication, we impose a discrete information bottleneck using vector quantization.
Specifically, the encoder output is mapped to a learned codebook and replaced by its nearest codeword, producing a discrete index map.

\paragraph{Entropy coding with rANS.}
The discrete indices exhibit significant redundancy because different codewords occur with different frequencies.
We therefore apply lossless entropy coding to pack indices into a compact bitstream before transmission.
We adopt range asymmetric numeral systems (rANS)~\cite{duda2014asymmetricnumeralsystemsentropy}, which approaches the Shannon limit in practice while remaining computationally efficient.
At the receiver, entropy decoding recovers the index sequence exactly, ensuring that no information is lost beyond the quantization step.

\paragraph{Receiver-side side-information network.}
On the receiver side, we exploit the availability of the local feature $\mathbf{F}_i$ by extracting a conditioning representation through a Side-Information Encoder.
SIEncoder mirrors the lightweight design of the sender encoder and produces conditioning features aligned in spatial resolution with the decoded latent.
We similarly employ residual connections to preserve the original receiver feature content while allowing SIEncoder to learn task-adaptive conditioning signals.

\paragraph{Conditional decoder and fusion-ready reconstruction.}
Finally, the receiver reconstructs a fusion-ready collaborator representation using a conditional decoder.
The decoder takes as input the decoded quantized latent from the sender and the conditioning feature from SIEncoder, concatenates them along the channel dimension, and applies a small stack of convolutional and residual blocks to produce $\hat{\mathbf{F}}_{j\rightarrow i}$.
This conditional reconstruction encourages the transmitted indices to capture information complementary to the receiver-local feature, while allowing the decoder to fill in predictable content from side information.
In this way, the codec is guided toward transmitting primarily the innovation beyond what is already available at the receiver, consistent with the DSC formulation.

\subsection{Training Objective}
\label{subsec:training_objective}

We train the overall framework end-to-end using downstream task supervision, augmented by a reconstruction objective that stabilizes the learned codec.
The overall loss is
\begin{equation}
\mathcal{L}=\mathcal{L}_{task}+\lambda\,\mathcal{L}_{rec},
\end{equation}
where $\mathcal{L}_{task}$ is the standard detection loss computed from the final prediction $\hat{\mathbf{y}}_i$.

\paragraph{Reconstruction loss.}
We regularize conditional communication by encouraging the reconstructed collaborator representation to match the transmitted feature:
\begin{equation}
\mathcal{L}_{rec}=\left\|\hat{\mathbf{F}}_{j\rightarrow i}-\tilde{\mathbf{F}}_j\right\|_2^2.
\end{equation}
In addition to the feature-level reconstruction term, $\mathcal{L}_{rec}$ also includes the standard vector-quantization regularizers (codebook update and commitment) returned by the quantizer, which empirically stabilizes learning of the discrete bottleneck.

\paragraph{Straight-through estimator.}
The discrete codeword assignment in vector quantization is non-differentiable.
We therefore adopt a straight-through estimator~\cite{van2017neural}: in the forward pass, latent vectors are replaced by their assigned codewords, while in the backward pass, gradients are passed through the quantization operation as an identity mapping.

\paragraph{Communication control.}
We do not explicitly optimize a rate term with a learned entropy model.
Instead, communication is primarily controlled by the discrete bottleneck configuration, and the realized payload is measured by the entropy-coded bitstream length in bytes.

\definecolor{upc}{RGB}{46,117,71}
\definecolor{downc}{RGB}{192,80,77}

\begin{table*}[t]
\centering
% \small
\setlength{\tabcolsep}{2.5pt}
\caption{\textbf{Overall performance and communication cost.}
We report mAP at IoU thresholds 0.3/0.5/0.7 on DAIR-V2X, OPV2V, and V2X-Real.
BD denotes the average per-link payload.
For Baseline+DCC, the first row is the original baseline, and the second row (colored) reports the result after inserting our DCC module while keeping the backbone and fusion head unchanged.
\textcolor{upc}{Green} indicates an improvement over the baseline, while \textcolor{downc}{Red} indicates a degradation.}

\label{tab:main}

\begin{tabular}{lcccccccccccc}
\toprule
Method & \multicolumn{4}{c}{DAIR-V2X} & \multicolumn{4}{c}{OPV2V} & \multicolumn{4}{c}{V2X-Real} \\
\cmidrule(lr){2-5} \cmidrule(lr){6-9} \cmidrule(lr){10-13}
& mAP@0.3 & mAP@0.5 & mAP@0.7 & BD & mAP@0.3 & mAP@0.5 & mAP@0.7 & BD & mAP@0.3 & mAP@0.5 & mAP@0.7 & BD \\
\midrule
\makecell{AttFuse\\+DCC} & \makecell{0.7132\\\textcolor{upc}{0.7734}} & \makecell{0.6520\\\textcolor{upc}{0.7288}} & \makecell{0.5077\\\textcolor{upc}
{0.5759}} & \makecell{8.00\,MB\\15.67\,KB} & \makecell{0.9298\\\textcolor{upc}{0.9394}} & \makecell{0.8986\\\textcolor{upc}{0.9327}} & \makecell{0.7706\\
\textcolor{upc}{0.8228}} & \makecell{16.00\,MB\\33.38\,KB} & \makecell{0.4756\\\textcolor{upc}{0.5116}} & \makecell{0.3907\\\textcolor{upc}{0.4058}}
& \makecell{0.2348\\\textcolor{upc}{0.2445}} & \makecell{8.59\,MB\\20.67\,KB} \\
\midrule
\makecell{CoAlign\\+DCC} & \makecell{0.7779\\\textcolor{upc}{0.7847}} & \makecell{0.7322\\\textcolor{upc}{0.7437}} & \makecell{0.5802\\\textcolor{upc}
{\underline{0.6012}}} & \makecell{8.00\,MB\\9.63\,KB} & \makecell{0.9498\\\textcolor{upc}{0.9551}} & \makecell{0.9442\\\textcolor{upc}{0.9503}} &
\makecell{0.8989\\\textcolor{upc}{0.9044}} & \makecell{16.00\,MB\\23.55\,KB} & \makecell{0.4849\\\textcolor{upc}{0.5006}} & \makecell{0.3964\\
\textcolor{upc}{0.4124}} & \makecell{0.2355\\\textcolor{upc}{\underline{0.2659}}} & \makecell{8.59\,MB\\15.67\,KB} \\
\midrule
\makecell{CoBEVT\\+DCC} & \makecell{0.7245\\\textcolor{upc}{0.7529}} & \makecell{0.6293\\\textcolor{upc}{0.6776}} & \makecell{0.3996\\\textcolor{upc}
{0.4559}} & \makecell{8.00\,MB\\10.96\,KB} & \makecell{0.9540\\\textcolor{downc}{0.9532}} & \makecell{0.9184\\\textcolor{upc}{0.9418}} &
\makecell{0.7680\\\textcolor{upc}{0.8482}} & \makecell{16.00\,MB\\17.20\,KB} & \makecell{0.4581\\\textcolor{upc}{0.4604}} & \makecell{0.3575\\
\textcolor{upc}{0.3921}} & \makecell{0.1559\\\textcolor{upc}{0.2356}} & \makecell{8.59\,MB\\16.32\,KB} \\
\midrule
\makecell{Disco\\+DCC} & \makecell{0.7096\\\textcolor{upc}{0.7674}} & \makecell{0.6487\\\textcolor{upc}{0.7197}} & \makecell{0.5062\\\textcolor{upc}
{0.5626}} & \makecell{8.00\,MB\\16.49\,KB} & \makecell{0.9175\\\textcolor{upc}{0.9333}} & \makecell{0.8765\\\textcolor{upc}{0.9308}} &
\makecell{0.7373\\\textcolor{upc}{0.8398}} & \makecell{16.00\,MB\\24.68\,KB} & \makecell{0.4811\\\textcolor{upc}{0.4906}} & \makecell{0.3925\\
\textcolor{upc}{0.4071}} & \makecell{0.2335\\\textcolor{upc}{0.2482}} & \makecell{8.59\,MB\\20.89\,KB} \\
\midrule
\makecell{F-Cooper\\+DCC} & \makecell{0.6731\\\textcolor{upc}{0.7807}} & \makecell{0.5666\\\textcolor{upc}{0.7366}} & \makecell{0.3877\\\textcolor{upc}
{0.5859}} & \makecell{8.00\,MB\\16.38\,KB} & \makecell{0.9126\\\textcolor{upc}{0.9447}} & \makecell{0.8253\\\textcolor{upc}{0.9378}} & \makecell{0.6180\\
\textcolor{upc}{0.8585}} & \makecell{16.00\,MB\\27.03\,KB} & \makecell{0.4365\\\textcolor{upc}{0.5147}} & \makecell{0.3238\\\textcolor{upc}{0.4203}} &
\makecell{0.1312\\\textcolor{upc}{0.2527}} & \makecell{8.59\,MB\\20.38\,KB} \\
\midrule
\makecell{V2X-ViT\\+DCC} & \makecell{0.7527\\\textcolor{upc}{0.7668}} & \makecell{0.6859\\\textcolor{upc}{0.7011}} & \makecell{0.4996\\\textcolor{upc}
{0.5059}} & \makecell{8.00\,MB\\11.98\,KB} & \makecell{0.9517\\\textcolor{upc}{\underline{0.9613}}} & \makecell{0.9300\\\textcolor{upc}{0.9471}} &
\makecell{0.8180\\\textcolor{upc}{0.8424}} & \makecell{16.00\,MB\\30.11\,KB} & \makecell{0.4960\\\textcolor{downc}{0.4837}} & \makecell{0.4002\\
\textcolor{downc}{0.3958}} & \makecell{0.2405\\\textcolor{upc}{0.2494}} & \makecell{8.59\,MB\\19.42\,KB} \\
\midrule
Where2comm & \underline{0.7933} & \underline{0.7484} & 0.5829 & 8.00\,MB & 0.9597 & \underline{0.9543} & \underline{0.9167} & 16.00\,MB & 0.5159 & 0.4259
& 0.2447 & 8.59\,MB \\
\midrule
Codefilling & 0.7900 & 0.7455 & 0.5919 & 28.06\,KB & 0.9562 & 0.9523 & 0.9105 & 60.52\,KB & \underline{0.5210} & \underline{0.4375} & 0.2553 & 29.39\,KB
\\
\midrule
Ours & \textbf{0.8004} & \textbf{0.7576} & \textbf{0.6028} & 10.04\,KB & \textbf{0.9720} & \textbf{0.9689} & \textbf{0.9322} & 27.44\,KB &
\textbf{0.5404} & \textbf{0.4489} & \textbf{0.2905} & 18.02\,KB \\
\bottomrule
\end{tabular}

\end{table*}

\section{Experiments}
\label{sec:experiments}

\subsection{Datasets and Experimental Settings}
\label{subsec:datasets_setup}
We evaluate our method on three cooperative 3D detection benchmarks spanning both simulation and real-world deployments: OPV2V~\cite{xu2022opv2v}, DAIR-V2X~\cite{yu2022dair}, and V2X-Real~\cite{xiang2024v2xreal}. 
Following standard practice, we report mean Average Precision (mAP) under multiple IoU thresholds (0.3/0.5/0.7). 
Communication overhead is measured by the average per-link message payload, i.e., the realized byte length of $m_{j\rightarrow i}$. 

OPV2V~\cite{xu2022opv2v} is a large-scale simulated benchmark dedicated to Vehicle-to-Vehicle (V2V) collaborative perception. Co-simulated using OpenCDA~\cite{xu2021opencda} and CARLA~\cite{dosovitskiy2017carla}, it provides high-fidelity 3D LiDAR point clouds and camera data sourced from diverse traffic scenarios. DAIR-V2X~\cite{yu2022dair} represents the first large-scale real-world dataset for V2X research, focusing on Vehicle-to-Infrastructure (V2I) scenarios. Data is captured from collaborating vehicles and roadside infrastructure units (RSUs) at various real urban intersections. V2X-Real~\cite{xiang2024v2xreal} is another challenging real-world dataset, specifically tailored for comprehensive Vehicle-to-Everything (V2X) cooperative perception under complex urban environments.

We use PointPillars~\cite{lang2019pointpillars} as the shared BEV backbone for all agents to ensure fair comparisons.
We train the whole framework end-to-end with Adam for 30 epochs, with an initial learning rate of $2\times10^{-3}$ and cosine annealing.
Unless otherwise specified, the DCC module uses a VQ codebook size $K=64$ with embedding dimension $D=64$. All runs are conducted on NVIDIA RTX 4090 GPUs.

We compare against representative intermediate-fusion collaborative perception methods, including strong communication-aware baselines such as Where2comm~\cite{hu2022where2comm} and Codefilling~\cite{hu2024communication}, as well as commonly adopted fusion pipelines (AttFuse~\cite{xu2022opv2v}, CoAlign~\cite{lu2023coalign}, CoBEVT~\cite{xu2023cobevt}, Disco~\cite{li2021disconet}, F-Cooper~\cite{chen2019fcooper}, and V2X-ViT~\cite{xu2022v2x}).
To further examine generality, we also report non-specialized communication-efficient baseline+DCC variants by integrating our DCC module into these pipelines without modifying their backbones or fusion heads.

\subsection{Quantitative Results}
\label{subsec:quant_results}

\begin{figure*}[t]
  \centering
    \subfloat[DAIR-V2X]{\includegraphics[width=\textwidth]{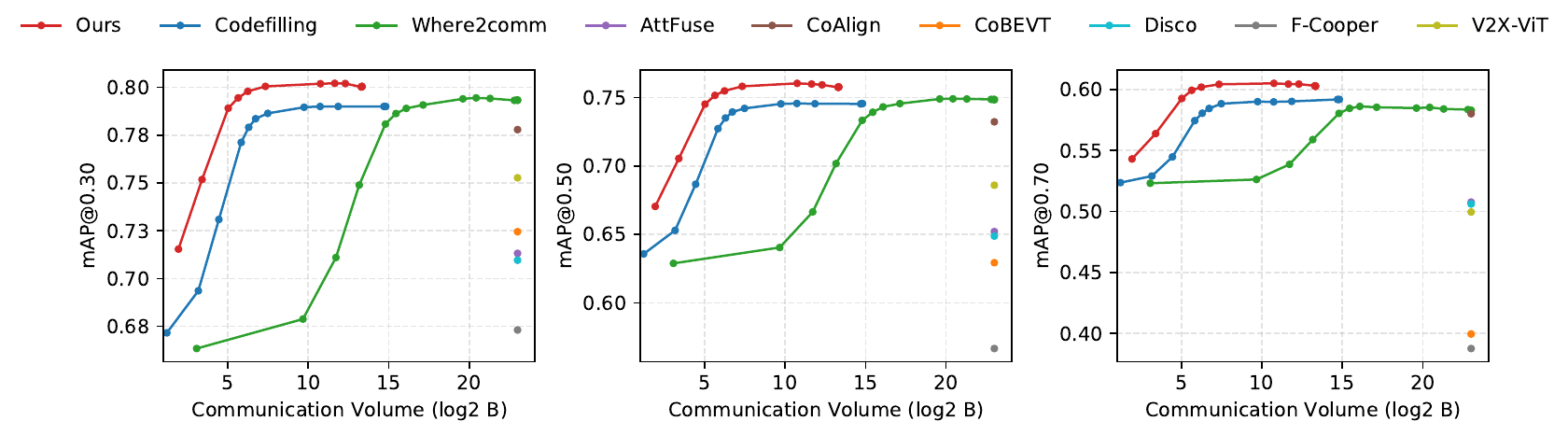}}\par\vspace{0.5mm}  
  \subfloat[OPV2V]{\includegraphics[width=\textwidth]{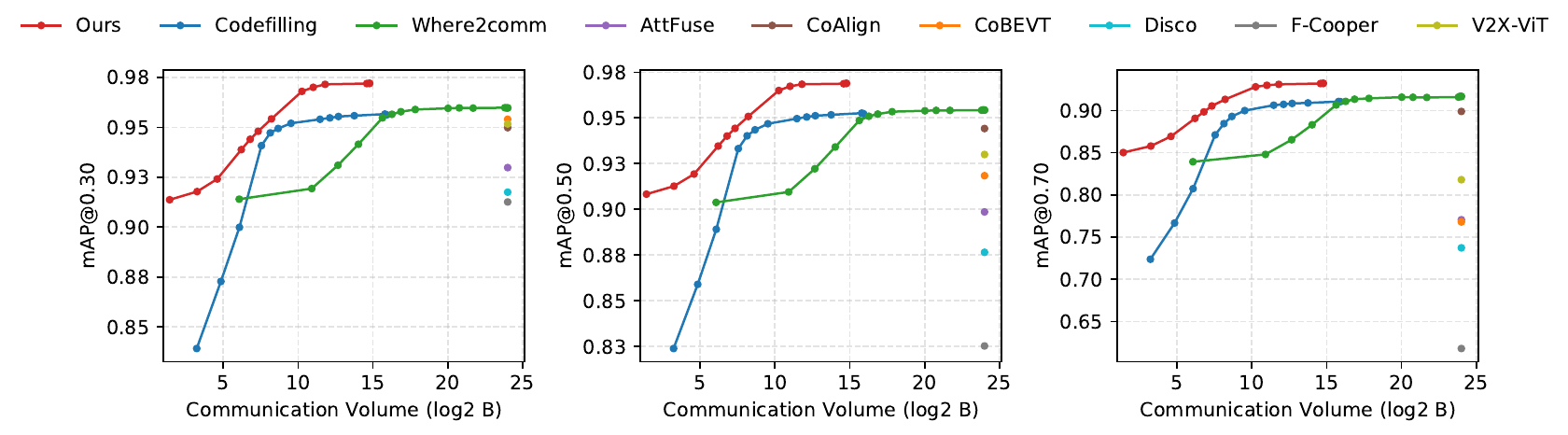}}\par\vspace{0.5mm}
  \subfloat[V2X-Real]{\includegraphics[width=\textwidth]{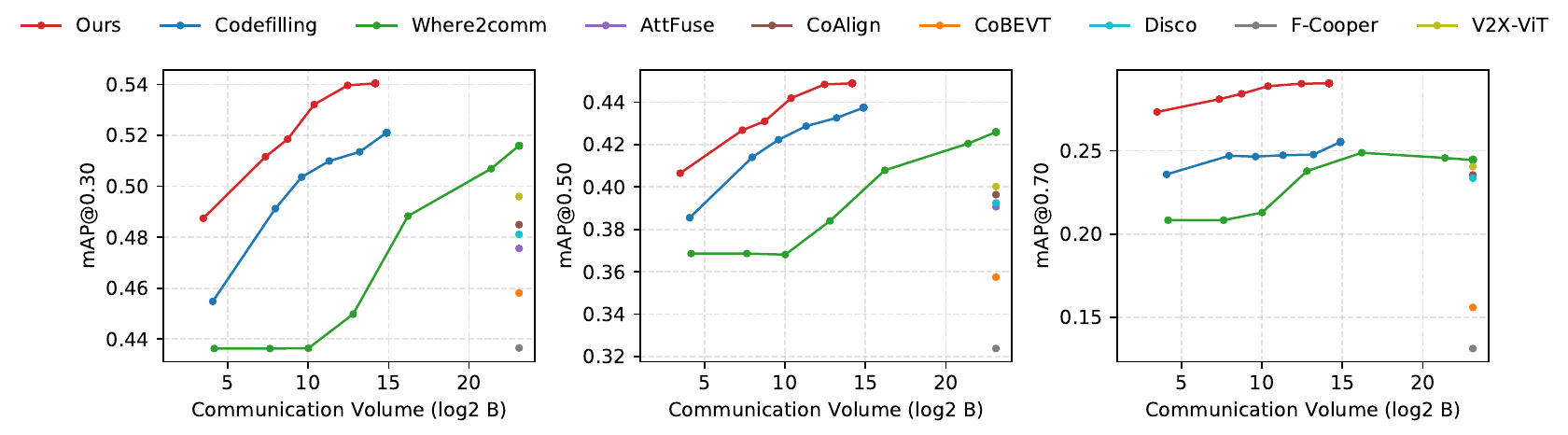}}
    \caption{\textbf{Communication--performance curves under different bandwidth budgets.}
    The x-axis shows the average per-link payload measured in bytes and plotted in $\log_2$ scale.
    We vary the effective communication rate by adjusting the pruning ratio, and report detection performance under the corresponding bandwidth.
    V2X-DSC achieves a favorable trade-off and remains on the Pareto frontier, particularly in the low-bandwidth regime.}

  \label{fig:comm_bw_all}
\end{figure*}

\begin{figure*}[t]
\centering
\subfloat[Noise Comparison]{\includegraphics[width=0.49\textwidth]{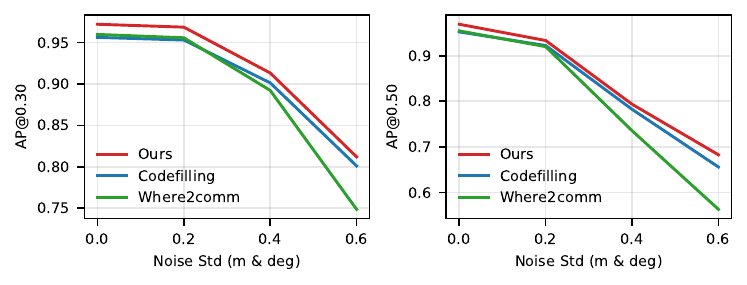}}
\hfill
\subfloat[Delay Comparison]{\includegraphics[width=0.49\textwidth]{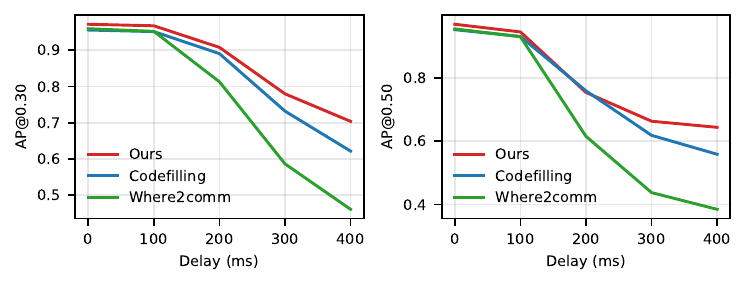}}
\caption{\textbf{Robustness analysis on the OPV2V dataset.} 
(a) Impact of Localization Noise: We evaluate detection performance under increasing Gaussian noise added to relative poses. V2X-DSC exhibits relatively stable performance compared to baselines as noise intensity increases.
(b) Impact of Communication Latency: Performance comparison under varying frame delays. Our method exhibits gradual performance degradation rather than a steep decline as latency increases.}
\label{fig:robustness_noise_delay}
\end{figure*}

\paragraph{Main comparison.}
Table~\ref{tab:main} summarizes the detection accuracy and communication cost on DAIR-V2X, OPV2V, and V2X-Real.
Overall, V2X-DSC consistently achieves a strong accuracy--bandwidth trade-off: it attains state-of-the-art mAP while operating under KB-level per-link payloads.
Compared to MB-level intermediate feature sharing (e.g., Where2comm), our method reduces communication by orders of magnitude without sacrificing accuracy.
Compared to compact discrete baselines (e.g., Codefilling), V2X-DSC achieves higher accuracy at lower bandwidth, demonstrating the advantage of conditional decoding that exploits receiver-local context.

Our method achieves 0.8004/0.7576/0.6028 mAP at IoU $0.3/0.5/0.7$ on DAIR-V2X with only 10.04 KB per link.
On OPV2V, V2X-DSC reaches 0.9720/0.9689/0.9322 with 27.44 KB, improving over Codefilling while using substantially fewer bits.
On V2X-Real, we obtain 0.5404/0.4489/0.2905 with 18.02 KB, outperforming prior compact-feature baselines under complex real-world scenarios.

\paragraph{Plug-and-play gains on diverse fusion backbones.}
To further assess generality, we insert our DCC module into several representative collaborative perception frameworks and keep the backbone and fusion head unchanged.
As shown in Table~\ref{tab:main} (rows ``Baseline+DCC''), DCC provides consistent bandwidth reduction and often improves accuracy across different fusion baselines.
These results indicate that the benefit of conditional reconstruction is not tied to a particular fusion design, and that DCC can serve as a drop-in communication layer for intermediate-fusion systems.

\paragraph{Accuracy--bandwidth curves.}
Figure~\ref{fig:comm_bw_all} reports accuracy--bandwidth curves under varying bandwidth budgets.
The x-axis shows the average per-link payload measured in bytes and plotted in $\log_2$ scale.
Across datasets, V2X-DSC remains on the Pareto frontier, with the most pronounced advantage in the low-bandwidth regime where standard compression or tokenization typically collapses task-relevant information.
This trend supports our central claim: by decoding with receiver-local context, DCC can preserve complementary cues even when the transmitted representation is extremely compact.

\subsection{Robustness to Pose Noise and Communication Delay}
\label{subsec:robustness}

Real-world V2X perception is inevitably affected by localization errors and asynchronous communication.
Figure~\ref{fig:robustness_noise_delay} evaluates robustness under increasing pose perturbations and communication delay.
Overall, V2X-DSC maintains competitive performance as noise and delay increase, indicating that the reconstructed features remain fusion-friendly under imperfect system conditions.

\subsection{Mechanistic Discussion}
\label{subsec:analysis}

We provide a brief interpretation of why the proposed DSC-guided conditional codec is effective under tight bandwidth and why it can often improve detection accuracy in practice.
From the DSC viewpoint, the receiver-local feature $\mathbf{F}_i$ encodes rich contextual information that is statistically correlated with the collaborator feature $\mathbf{F}_j$.
Therefore, transmitting a compact innovation code can be sufficient: predictable content can be recovered from local context, while transmitted bits are concentrated on complementary cues.

Empirically, we observe that V2X-DSC can match or exceed baselines even at extremely low payloads.
We hypothesize two contributing factors.
First, the side-information branch provides a strong prior during reconstruction, which helps correct quantization noise and suppress harmful inconsistencies introduced by aggressive compression.
Second, the discrete bottleneck (vector quantization) acts as an implicit regularizer: it filters out unstable feature variations and encourages a more compact, fusion-consistent representation manifold, which can reduce overfitting to sender-specific noise patterns.
These effects are particularly beneficial in low-bandwidth settings, where naive independent compression tends to discard geometrically critical information.

\section{Conclusion}
In this paper, we revisited bandwidth-constrained multi-agent collaborative perception from the perspective of distributed source coding.
Motivated by the strong inter-agent correlation in intermediate BEV representations, we proposed V2X-DSC, an intermediate-fusion framework equipped with a DSC-guided Conditional Codec.
The codec follows an asymmetric design: the sender independently encodes collaborator features into compact messages, while the receiver performs side-information-conditioned reconstruction using its local BEV feature, enabling innovation-centric communication under tight bandwidth budgets.

Extensive experiments on OPV2V, DAIR-V2X, and V2X-Real demonstrate that V2X-DSC achieves a strong accuracy--bandwidth trade-off, establishing a new Pareto frontier in the low-bandwidth regime and integrating seamlessly with multiple fusion backbones.
We hope this work encourages further exploration of information-theoretic principles for practical communication design in collaborative perception.

% \section*{Acknowledgments}
% This should be a simple paragraph before the References to thank those individuals and institutions who have supported your work on this article.

% {\appendix[Proof of the Zonklar Equations]
% Use $\backslash${\tt{appendix}} if you have a single appendix:
% Do not use $\backslash${\tt{section}} anymore after $\backslash${\tt{appendix}}, only $\backslash${\tt{section*}}.
% If you have multiple appendixes use $\backslash${\tt{appendices}} then use $\backslash${\tt{section}} to start each appendix.
% You must declare a $\backslash${\tt{section}} before using any $\backslash${\tt{subsection}} or using $\backslash${\tt{label}} ($\backslash${\tt{appendices}} by itself
%  starts a section numbered zero.)}


\bibliographystyle{IEEEtran}
\bibliography{bib/references}

% \newpage

 

% \end{IEEEbiography}



\vfill

\end{document}